%% file: main.tex
\documentclass[sigconf]{acmart}

\renewcommand\footnotetextcopyrightpermission[1]{}

\setcopyright{none}

\usepackage{graphicx}
\usepackage{booktabs} 
\usepackage[utf8]{inputenc} 
\usepackage[T1]{fontenc}    
\usepackage{booktabs}       
\usepackage{amsfonts}       
\usepackage{amsmath}
\usepackage{amssymb}
\usepackage{nicefrac}       
\usepackage{microtype}      
\usepackage{color}
\usepackage{subcaption}
\usepackage{enumitem}
\usepackage{xspace}
\usepackage[export]{adjustbox}
\usepackage{natbib}
\usepackage{cancel}
\usepackage[printwatermark]{xwatermark}
\usepackage[normalem]{ulem}
\setcitestyle{square,numbers,sort}

\graphicspath{{figures/}}

\newif{\ifhidecomments}
\hidecommentsfalse
\ifhidecomments
    \newcommand{\chenhao}[1]{}
    \newcommand{\amit}[1]{}
    \newcommand{\ram}[1]{}
\else
    \newcommand{\chenhao}[1]{\textcolor{blue}{[#1 ---\textsc{ct}]}}
    \newcommand{\amit}[1]{\textcolor{blue}{[#1 ---\textsc{am}]}}
     \newcommand{\ram}[1]{\textcolor{red}{#1}}
\fi

\newcommand{\para}[1]{\noindent {\bf #1}\xspace}
\newcommand{\figref}[1]{Figure~\ref{#1}\xspace}
\newcommand{\secref}[1]{Section~\ref{#1}\xspace}

\copyrightyear{2020} 
\acmYear{2020} 
\setcopyright{acmcopyright}
\acmConference[FAT* '20]{Conference on Fairness, Accountability, and Transparency}{January 27--30, 2020}{Barcelona, Spain}
\acmBooktitle{Conference on Fairness, Accountability, and Transparency (FAT* '20), January 27--30, 2020, Barcelona, Spain}
\acmPrice{15.00}
\acmDOI{10.1145/3351095.3372850}
\acmISBN{978-1-4503-6936-7/20/02}

\author{Ramaravind K. Mothilal}
\affiliation{
  \institution{Microsoft Research India}
}
\email{t-rakom@microsoft.com}

\author{Amit Sharma}
\affiliation{
  \institution{Microsoft Research India}
}
\email{amshar@microsoft.com}

\author{Chenhao Tan}
\affiliation{
  \institution{University of Colorado Boulder}
}
\email{chenhao.tan@colorado.edu}

\ccsdesc[500]{Applied computing~Law, social and behavioral sciences}

\newcommand{\vect}[1]{\boldsymbol{#1}}
\newcommand{\vecx}{\vect{x}}
\newcommand{\counterfactuals}{\mathcal{C}}
\newcommand{\counterfactual}{\vect{c}}

\newcommand{\Comb}[2]{C_{#1}^{#2}}
\newcommand{\classifier}{f}
\DeclareMathOperator*{\argmin}{arg\,min}

\newcommand{\compas}{\texttt{COMPAS}\xspace}
\newcommand{\adult}{\texttt{Adult-Income}\xspace}
\newcommand{\german}{\texttt{German-Credit}\xspace}
\newcommand{\lclub}{\texttt{LendingClub}\xspace}
\newcommand{\singleCF}{\texttt{SingleCF}\xspace}
\newcommand{\randomCF}{\texttt{RandomInitCF}\xspace}
\newcommand{\nodiverseCF}{\texttt{NoDiversityCF}\xspace}
\newcommand{\diverseCF}{\texttt{DiverseCF}\xspace}
\newcommand{\diverseCFsp}{\texttt{DiverseCF-Sparse}\xspace}
\newcommand{\lime}{\texttt{LIME}\xspace}
\newcommand{\chrisCF}{\texttt{MixedIntegerCF}\xspace}

\begin{document}

\title{Explaining Machine Learning Classifiers through Diverse Counterfactual Explanations}
\renewcommand{\shorttitle}{Diverse Counterfactual Explanations}

\begin{abstract}

Post-hoc explanations of machine learning models are crucial for people to understand and act on algorithmic predictions.
An intriguing class of explanations is through \emph{counterfactuals}, hypothetical examples that show people how to obtain a different prediction.
We posit that effective counterfactual explanations should satisfy two properties: \emph{feasibility} of the counterfactual actions given user context and constraints, and \emph{diversity} among the counterfactuals presented. 
To this end, we propose a framework for generating and evaluating a diverse set of counterfactual explanations based on 
determinantal point processes.  
To evaluate the actionability of counterfactuals, we provide metrics that  enable  comparison of counterfactual-based methods to other local explanation methods. 
We further address necessary tradeoffs and point to causal implications in optimizing for counterfactuals. 
Our experiments on four real-world datasets show that our framework can generate a set of counterfactuals that are diverse and well approximate local decision boundaries, outperforming prior approaches to generating diverse counterfactuals. We provide an implementation of the framework at {\color{blue}\url{https://github.com/microsoft/DiCE}}.

\end{abstract}

\maketitle

\input{intro}

\input{related}

\input{model}

\input{exp}

\input{results}

\input{causal}

\input{conclusion}

\vspace{1em}
\para{Acknowledgments}.
We thank Brian Lubars for his insightful comments and Chris Russel for providing assistance in running the diverse CF generation on linear models.
This work was supported in part by NSF grant IIS-1927322.

\bibliographystyle{ACM-Reference-Format}
\interlinepenalty=10000
\bibliography{cf-explain2}

\input{supp}

\end{document}

%% file: intro.tex
\section{Introduction}

Consider a person who applied for a loan and was rejected by the loan distribution algorithm of a financial company. Typically, the company may provide an explanation on why the loan was rejected, for example, due to ``poor credit history''. However, such an explanation does not help the person decide \textit{what they should do next} to improve their chances of being approved in the future. Critically, the most important feature may not be enough to flip the decision of the algorithm, and in practice, may not even be changeable such as gender and race.
Thus, it is equally  important to show decision outcomes from the algorithm 
with \emph{actionable} alternative profiles, to help people understand what they could have done to change their loan decision. 
Similar to the loan example, this argument is valid for a range of scenarios involving decision-making on an individual's outcome, such as deciding admission to a university~\cite{waters2014grade}, screening job applicants \cite{rockoff2011can}, disbursing government aid \cite{andini2017targeting,athey2017beyond}, and identifying people at high risk of a future disease \cite{dai2015prediction}. In all these cases, knowing reasons for a bad outcome is not enough; it is important to know what to do to obtain a better outcome in the future (assuming that the algorithm remains relatively static). 

\emph{Counterfactual} explanations~\cite{wachter2017counterfactual} provide this information, by showing feature-perturbed versions of the same person who would have received the loan, e.g., ``you would have received the loan if your income was higher by $\$10,000$''. In other words, they provide ``what-if'' explanations for model output. Unlike explanation methods that depend on approximating the classifier's decision boundary~\cite{ribeiro2016should},
counterfactual (CF) explanations have the advantage that they are always truthful w.r.t. the underlying model by giving direct outputs of the algorithm.  Moreover, counterfactual examples may also be human-interpretable~\cite{wachter2017counterfactual} by allowing users to explore ``what-if'' scenarios,
similar to how children learn through counterfactual examples~\cite{weisberg2013pretense,beck2009relating,buchsbaum2012power}.

However, it is difficult to generate CF examples 
that  are \emph{actionable} for a person's situation.
Continuing our loan decision example, a CF explanation may  suggest to ``change your house rent'', but it does not say much about  alternative counterfactuals, or consider the relative ease between different changes a person may need to make. Like  any example-based decision support system~\cite{kim2016examples}, 
we need a \emph{set} of counterfactual examples to help a person interpret a complex machine learning model. Ideally, these examples 
should balance between a wide range of suggested changes (\emph{diversity}), and the relative ease of adopting those changes (\emph{proximity} to the original input), and also follow the causal laws of human society, e.g., one can hardly lower their educational degree or change their race. 

Indeed, \citet{russell2019efficient} recognizes the importance of diversity and proposes an approach for linear machine learning classifiers based on integer programming.
In this work, we propose a method that generates sets of diverse counterfactual examples for any differentiable machine learning classifier. Extending \citet{wachter2017counterfactual}, we construct an optimization problem that considers the diversity of the generated CF examples, in addition to proximity to the original input. Solving the optimization problem requires considering the tradeoff between diversity and proximity, and the tradeoff between continuous and categorical features which may differ in their relative scale and ease of change. We provide a general solution to this optimization problem that can generate any number of CF examples for a given input. 
To facilitate actionability, our solution is flexible enough to support user-provided inputs based on domain knowledge, such as custom weights for individual features or constraints on perturbation of features.

Further, we provide quantitative evaluation metrics for evaluating any set of counterfactual examples. Due to their inherent subjectivity, CF examples are hard to evaluate.
While we cannot replace behavioral experiments, we propose  metrics that can help 
in 
fine-tuning parameters of the proposed solution to achieve desired properties of validity, diversity, and proximity. 
We also propose a second evaluation metric that approximates 
a behavioral experiment on 
whether people can  \emph{understand} a ML model's decision given a set of CF examples, assuming that people would rationally extrapolate from the CF examples and ``guess'' the local decision boundary of an ML model.

We evaluate our method on explaining ML models trained on four datasets: COMPAS for bail decision~\cite{propublica-story}, Adult-Income for income prediction~\cite{adult}, German-Credit for assessing credit risk~\cite{german}, and a dataset from Lending Club for loan decisions~\cite{lending-data}.  Compared to prior CF generation methods, our proposed solution generates CF examples with substantially higher diversity for 
these datasets. Moreover, 
a simple 1-nearest neighbor model trained on the generated CF examples obtains comparable accuracy on locally approximating the original ML model to methods like LIME~\cite{ribeiro2016should}, which are directly optimized for estimating the local decision boundary. Notably, our method obtains higher F1 score on predicting instances in the counterfactual outcome class than LIME in most configurations, especially for Adult-Income and COMPAS datasets wherein both precision and recall are higher. Qualitative inspection of the generated CF examples illustrates their potential utility for making informed decisions. Additionally, CF explanations can expose biases in the original ML model, as we see when some of the generated explanations suggest changes in sensitive attributes like race or gender.  The last example illustrates the broad applicability of CF explanations: they are not just useful to an end-user, but can be equally useful to model builders for debugging biases, and for fairness evaluators to discover such biases and other model properties.

Still, CF explanations, as generated,  suffer from lack of any causal knowledge about the input features that they modify. Features do not exist in a vacuum; they come from a data-generating process which constrains their modification. Thus, perturbing  each input feature independently can lead to infeasible examples, such as suggesting someone to obtain a higher 
degree but reduce their age.
To ensure feasibility, we propose a filtering approach on the generated CF examples based on causal constraints.

To summarize, our work makes the following contributions:
\begin{itemize}[itemsep=0pt,leftmargin=*,topsep=2pt]
    \item We propose diversity as an important component for actionable counterfactuals and build a general optimization framework that exposes the importance of necessary tradeoffs, causal implications, and optimization issues in generating counterfactuals.  
    \item We propose a quantitative evaluation framework for counterfactuals that allows fine-tuning of the proposed method for a particular scenario and enables comparison of CF-based methods to other local explanation methods such as LIME.
    \item Finally, we  
    demonstrate the effectiveness of our framework
    through empirical experiments on multiple datasets 
    and provide an open-source implementation 
    at {\color{blue} \url{https://github.com/microsoft/DiCE}}. 
\end{itemize}

%% file: related.tex
\section{Background \& Related work}

Explanations are critical for machine learning, especially as machine learning-based systems are being used to inform decisions in societally critical domains such as finance, healthcare, education, and criminal justice. Since many machine learning algorithms 
are black boxes to end users and do not provide guarantees on input-output relationship, explanations serve a useful role to inspect these models.
Besides helping to debug ML models, explanations are hypothesized to improve the interpretability and trustworthiness of algorithmic decisions and enhance human decision making
\cite{doshi2017towards,lipton2016mythos,tomsett2018interpretable,kusner2017counterfactual}.  
Below we focus on 
approaches that provide post-hoc explanations of machine learning models and discuss why diversity should be an important component for counterfactual explanations.
There is also 
an important line of work that focuses on developing intelligible models by assuming that simple models such as linear models or decision trees are interpretable \cite{lakkaraju2016interpretable,lou2013accurate,lou2012intelligible,caruana2015intelligible}.

\subsection{Explanation through Feature Importance}

An important approach to post-hoc explanations is to determine feature importance for a particular prediction through \emph{local} approximation.
\citet{ribeiro2016should} propose a feature-based approach, LIME, that fits a sparse linear model to approximate non-linear models locally. \citet{guidotti2018lore} extend this approach by fitting a decision-tree classifier to approximate the non-linear model and then tracing the decision-tree paths to generate explanations.     
Similarly, \citet{lundberg2017unified} present a unified framework that assigns each feature an importance value for a particular prediction. 
Such explanations, however, ``lie'' about the machine learning models. There is an inherent tradeoff between truthfulness about the model and human interpretability when explaining a complex model, and so explanation methods that use proxy models inevitably approximate the true model to varying degrees. 
Similarly, \emph{global} explanations can be generated by approximating the true surface with a simpler surrogate model and using the simpler model to derive explanations \cite{ribeiro2016should,craven1996extracting}. 
A major problem with these approaches is that since the explanations are sourced from simpler surrogates, there is no guarantee that they are faithful to the original model.

\subsection{Explanation through Visualization}
Similar to identifying feature importance, visualizing the decision of a model is a common technique for explaining model predictions.
Such visualizations are commonly used in the computer vision community, ranging from highlighting certain parts of an image to activations in convolutional neural networks \citep{zhou2018interpretable,zeiler2014visualizing,mahendran2015understanding}.
However, these visualizations can be difficult to interpret in scenarios that are not inherently visual such as recidivism prediction and loan approvals, which are the cases that our work focuses on.

\subsection{Explanation through Examples}

The most relevant class of explanations to our approach is through examples.
An example-based explanation framework is MMD-critic proposed by \citet{kim2016examples}, which selects both prototypes and criticisms from the original data points.
More recently, counterfactual explanations are proposed as a way to provide alternative perturbations that would have changed the prediction of a model.
In other words, given an input feature $\vecx$ and the corresponding output by a ML model $f$, 
a {\em counterfactual explanation} is a perturbation of the input to generate a different output $y$ by the same algorithm. 
Specifically, \citet{wachter2017counterfactual} propose the following formulation:
\begin{align}
\counterfactual = \argmin_{\counterfactual} & \operatorname{yloss}(\classifier(\counterfactual), y) 
                                 +  | \vecx - \counterfactual |, \label{eq:dist}
\end{align}
\noindent where the first part ($\operatorname{yloss}$) pushes the counterfactual $\counterfactual\ $ towards a different prediction than the original instance, and the second part keeps the counterfactual close to the original instance.

Extending their work, we provide a method to construct a set of counterfactuals with diversity. In other domains of information search such as search engines and recommendation systems, multiple studies \citep{ziegler2005improving,ekstrand2014user,kunaver2017diversity,sanderson2009else} show the benefits of presenting a diverse set of information items to a user. Our hypothesis is that diversity can be similarly beneficial when people are shown  counterfactual explanations. 
For linear models, a recent paper by \citet{russell2019efficient} develops an efficient algorithm to find diverse counterfactuals using integer programming.
In this work, we examine an alternative formulation that works for any differentiable model, investigate multiple practical issues on different datasets, and propose a general quantitative evaluation framework for diverse counterfactuals.

%% file: model.tex
\section{Counterfactual Generation Engine} 
\label{sec:cf-engine}

The input of our problem is a trained machine learning model, $f$, and an instance, $\vecx$. We would like to generate a set of $k$ counterfactual examples, $\{\counterfactual_1, \counterfactual_2, \ldots, \counterfactual_k\}$, such that they all lead to a different decision than $\vecx$. 
The instance ($\vecx$) and all CF examples ($\{\counterfactual_1, \counterfactual_2, \ldots, \counterfactual_k\}$) are \textit{d}-dimensional. Throughout the paper, we assume that the machine learning model is differentiable and static (does not change over time), and that the output is binary. Table~\ref{tab:terminology} summarizes the main terminologies used in the paper.

\begin{table}[t]
\centering
\small{
\begin{tabular}{p{1in}p{2.2in}}
\toprule
ML model ($f$) & The trained model obtained from the training data. \\
Original input ($\vecx$) & The feature vector associated with an instance of interest that receives an unfavorable decision from the ML model. \\
Original outcome & The prediction of the original input from the trained model, usually corresponding to the undesired class.\\
Original outcome class & The undesired class. \\
Counterfactual example ($\counterfactual_i$) & An instance (and its feature vector) close to the original input that would have received a favorable decision from the ML model. \\
CF class & The desired class. \\
\bottomrule
\end{tabular}}
\caption{Terminology used throughout the paper.}
\label{tab:terminology}
\end{table}

Our goal is to generate an actionable counterfactual set, that is, the user should be able to find CF examples that they can act upon. 
To do so, we need individual CF examples to be feasible with respect to the original input, but also need diversity among the generated counterfactuals to provide different ways of changing the outcome class.  Thus, we adapt diversity metrics to generate diverse counterfactuals that can offer users multiple options (\secref{sec:diversity}).
At the same time, we incorporate feasibility using the proximity constraint from \citet{wachter2017counterfactual} and introduce other user-defined constraints. 
Finally, we 
point out that counterfactual generation is a post-hoc procedure distinct from the standard machine learning setup, and discuss related practical issues (\secref{sec:practical}).

\subsection{Diversity and 
 Feasibility Constraints}
\label{sec:diversity}
Although diverse CF examples increase the chances that at least one example will be actionable for the user,  
examples may end up changing a large set of features, or maximize diversity by considering big changes from the original input.
This situation could be worsened when features are high-dimensional. 
We thus need a combination of diversity and feasibility, as we formulate below.

\para{Diversity via Determinantal Point Processes.}
We capture 
diversity by building on determinantal point processes (DPP), 
which has been adopted for solving subset selection problems with diversity constraints \citep{kulesza2012determinantal}.
We use the following metric based on the determinant of the kernel matrix given the counterfactuals:
\begin{align} \label{eqn:diversity-dpp}
dpp\_diversity = det({\bf K}),
\end{align}

\noindent where ${\bf K}_{i,j} = \frac{1}{1+dist(\counterfactual_i, \counterfactual_j)}$ and $dist(\counterfactual_i, \counterfactual_j)$ denotes a distance metric between the two counterfactual examples.  
In practice, to avoid ill-defined determinants, we add small random perturbations to 
the diagonal elements 
for computing the determinant.

\para{Proximity.} Intuitively, CF examples that are closest to the original input can be the most useful to a user. We quantify \emph{proximity} as the (negative) vector distance between the original input and CF example's features. This can be specified by a distance metric such as $\ell_1$-distance (optionally weighted by a user-provided custom weight for each feature). Proximity of a set of counterfactual examples is the mean proximity over the set.
\begin{align}
    Proximity :=  -\frac{1}{k} \sum_{i=1}^{k}  dist(\counterfactual_i, \vecx).
\end{align}

\para{Sparsity.}
Closely connected to proximity is the feasibility property of sparsity: how many features does a user need to change to transition to the counterfactual class. Intuitively, a counterfactual example will be more feasible if it makes changes to fewer number of features. Since this constraint is non-convex, we do not include it in the loss function but rather handle it through modifying the generated counterfactuals, as explained in Section~\ref{sec:practical}. 

\para{User constraints.}
A counterfactual example may be close in feature space, but may not be feasible due to real world constraints. Thus, it makes sense to allow the user to provide constraints on feature manipulation. They can be specified in two ways. First, as box constraints on feasible ranges for each feature, within which CF examples need to be searched. An example of such a constraint is: ``income cannot increase beyond 200,000''. Alternatively, a user may specify the variables that can be changed.

In general, feasibility is a broad issue that encompasses many facets.
We further examine a novel feasibility constraint derived from causal relationships in \secref{sec:causal}.

\subsection{Optimization} 
\label{sec:opt}

Based on the above definitions of diversity and proximity, we consider a combined loss function over all generated counterfactuals. 
\begin{align}
    \counterfactuals(\vecx) = \argmin_{\counterfactual_1, \ldots, \counterfactual_k} &  \frac{1}{k} \sum_{i=1}^{k} \operatorname{yloss}(\classifier(\counterfactual_i), y) +
                                  \frac{\lambda_1}{k} \sum_{i=1}^{k} dist(\counterfactual_i, \vecx)  \nonumber\\  
                                 & - \lambda_2 \operatorname{dpp\_diversity}(\counterfactual_1, \ldots, \counterfactual_k) \label{eq:diversity}
\end{align}
where $\counterfactual_i$ is a counterfactual example (CF), $k$ is the total number of CFs to be generated, $\classifier(.)$ is the ML model (a black box to end users), $yloss(.)$ is a metric that minimizes the distance between $f(.)$'s prediction for $\counterfactual_i$s and the desired outcome $y$ (usually 1 in our experiments),  $d$ is the total number of input features, $\vecx$ is the original input, and $\operatorname{dpp\_diversity}(.)$ is the diversity metric. $\lambda_1$ and $\lambda_2$ are 
hyperparameters that balance the three parts of the loss function.

\para{Implementation.} We optimize the above loss function using gradient descent. 
Ideally, we can achieve $f(\counterfactual_i)=y$ for every counterfactual,
but this may not always be possible because the objective is non-convex.
We run a maximum of 5,000 steps, or until the loss function converges and the generated counterfactual is valid (belongs to the desired class). 
We initialize all $\counterfactual_i$ randomly.

\subsection{Practical considerations}
\label{sec:practical}
Important practical considerations need to be made for such counterfactual algorithms to work in practice, since they involve multiple tradeoffs in choosing the final set. Here we describe four such considerations. 
While these considerations might seem trivial from a technical perspective, we believe that they are important for supporting user interaction with counterfactuals.

\para{Choice of yloss. }
An intuitive choice of $yloss$ may be $\ell_1$-loss ($|y-f(\counterfactual)|$) or $\ell_2$-loss. However, these loss functions penalize the distance of $f(\counterfactual)$ from the desired $y$, whereas a valid counterfactual only requires that $f(\counterfactual)$ be greater or lesser than f's threshold (typically 0.5), not necessarily the closest to desired $y$ (1 or 0). In fact, optimizing for $f(\counterfactual)$ to be close to either 0 or 1 encourages large changes to $\vecx$ towards the counterfactual class, which in turn make the generated counterfactual less feasible for a user.  Therefore, we use a hinge-loss function that ensures zero penalty as long as $f(\counterfactual)$ is above a fixed threshold above 0.5 when the desired class is 1 (and below a fixed threshold when the desired class is 0). Further, it imposes a penalty proportional to difference between $f(c)$ and 0.5 when the classifier is correct (but within the threshold), and a  heavier penalty when $f(c)$ does not indicate the desired  counterfactual class. Specifically, the hinge-loss is: 
$$hinge\_yloss = max(0, 1 - z * logit(f(\counterfactual))),$$

\noindent where $z$ is -1 when $y=0$ and 1 when $y=1$, and $logit(f(\counterfactual))$ is the unscaled output from the ML model (e.g.,  final logits that enter a softmax layer for making predictions in a neural network).

\para{Choice of distance function. }
For continuous features, we define $dist$ as the mean of feature-wise $\ell_1$ distances between the CF example and the original input. Since features can span different ranges, we divide each feature-wise distance by the median absolute deviation (MAD) of the feature's values in the training set, following \citet{wachter2017counterfactual}.  Deviation from the median provides a robust measure of the variability of a feature's values, and thus dividing by the MAD allows us to capture the relative prevalence of observing the feature at a particular value.
\begin{equation} \label{eqn:dist-cont}
 \operatorname{dist\_cont}(\counterfactual, \vecx) = \frac{1}{d_{cont}}\sum\limits_{p=1}^{d_{cont}} \frac{|\counterfactual^{p} - \vecx^{p}|}{MAD_p},
 \end{equation}
\noindent where $d_{cont}$  is the number of continuous variables and $MAD_p$ is the median absolute deviation for the $p$-th continuous variable.

For categorical features, however, it is unclear how to define a notion of distance. While there exist metrics based on the relative frequency of different categorical levels for a feature in available data~\cite{whatifurl}, they may not correspond to the \emph{difficulty} of changing a particular feature. For instance, irrespective of the relative ratio of different education levels (e.g., high school or bachelors), it is quite hard to obtain a new educational degree, compared to changes in other categorical features. We thus use a simpler metric that assigns a distance of 1 if the CF example's value for any categorical feature differs from the original input, otherwise it assigns zero. 
\begin{equation}\label{eqn:dist-cat}
\operatorname{dist\_cat}(\counterfactual, \vecx) = \frac{1}{d_{cat}}\sum\limits_{p=1}^{d_{cat}} I(\counterfactual^{p} \neq \vecx^{p}),
\end{equation}
\noindent where $d_{cat}$ is the number of categorical variables.

\para{Relative scale of features.}
In general, continuous features can have a wide range of possible values, while typical encoding for categorical features constrains them to a one-hot binary representation.
Since the scale of a feature highly influences how much it matters in our objective function, we believe that the ideal solution is to provide interactive interfaces to allow users to input their preferences across features.
As a sensible default, however, we transform all features to $[0,1]$. Continuous features are simply scaled between 0 and 1. For categorical features, we convert each feature using one-hot encoding and consider it as a continuous variable between 0 and 1.  
Also, to enforce the one-hot encoding in the learned counterfactuals, we add a regularization term with high penalty for each categorical feature to force its values for different levels to sum to 1. At the end of the optimization, we pick the level with maximum value for each categorical feature.

\para{Enhancing Sparsity.}
While our loss function minimizes the distance between the input and the generated counterfactuals, an ideal counterfactual 
needs to be sparse in the number of features it changes. 
To encourage sparsity in a generated counterfactual, we conduct a post-hoc operation where we restore the value of continuous features back to their values in $\vecx$ greedily until the predicted class $f(\counterfactual)$  changes. For this operation,  we consider all continuous features $\counterfactual^j$ whose difference from $\vecx^j$ is less than a chosen threshold.
Although an intuitive 
threshold is the median absolute distance (MAD),
the MAD can be fairly large for features with large variance. Therefore, for each feature, we choose the minimum of MAD and the bottom $10\%$ percentile of the absolute difference between non-identical values from the median.

\para{Hyperparameter choice.}
    Since counterfactual generation is a post-hoc step after training the ML model, it is not necessarily required that we use the same hyperparameter for every original input \citep{wachter2017counterfactual}.
    However, since hyperparameters can influence the generated counterfactuals, it seems problematic if users are given counterfactuals generated by different hyperparameters.\footnote{In general, whether the explanation algorithm should be uniform is a fundamental issue for providing post-hoc explanations of algorithmic decisions and it likely depends on the nature of such explanations.}
    In this work, therefore, we choose $\lambda_1=0.5$ and $\lambda_2=1$ based on a grid-search with different values and evaluating the diversity and proximity of generated CF examples.

%% file: exp.tex
\section{Evaluating counterfactuals}
Despite recent interest in counterfactual explanations \citep{russell2019efficient,wachter2017counterfactual}, the evaluations are typically only done in a qualitative fashion.
In this section, 
we present metrics for evaluating the quality of a set of counterfactual examples. 
As stated in Section~\ref{sec:cf-engine}, it is desirable that a method produces diverse and proximal examples and that it can generate valid counterfactual examples for all possible inputs. Ultimately, however, the examples should help a user in understanding the local decision boundary of the ML classifier.  
Thus, in addition to diversity and proximity, we propose a metric that approximates the notion of a user's understanding. We do so by constructing a secondary model based on the counterfactual examples that acts as a proxy of a user's understanding, and compare how well it can mimic the ML classifier's decision boundary.

Nevertheless, it is important to emphasize that CF examples are eventually evaluated by end users.
The goal of this work is to provide metrics that pave the way towards meaningful human subject experiments,
and we will offer further discussion in \secref{sec:conclusion}.

\subsection{Validity, Proximity, and Diversity}
First, we define 
quantitative metrics for validity, diversity, and proximity for a counterfactual set that can be used to evaluate any method for generating counterfactuals. We assume that a set $\mathcal{C}$ of $k$ counterfactual examples are generated for an original input.  

\para{Validity.}
Validity is simply the fraction of examples returned by a method that are actually \emph{counterfactuals}. That is, they correspond to a different outcome than the original input. Here we consider only unique examples because a method may generate multiple examples that are identical to each other.
$$\% \texttt{Valid-CFs} = \frac{|\{\text{unique instances in } \mathcal{C} \text{ s.t. } f(c) > 0.5\}|}{k}$$

\para{Proximity.}
\noindent We define distance-based proximity separately for continuous and categorical features. Using the definition of $dist$ metrics from Equations~\ref{eqn:dist-cont} and \ref{eqn:dist-cat}, we define proximity as:
\begin{align}
    \texttt{Continuous-Proximity}: -\frac{1}{k} \sum\limits_{i=1}^{k} \operatorname{dist\_cont}(\counterfactual_i, \vecx), \label{eq:cont-prox-eval} \\
    \texttt{Categorical-Proximity}:1 -\frac{1}{k} \sum\limits_{i=1}^{k} \operatorname{dist\_cat}(\counterfactual_i, \vecx), \label{eq:cat-prox-eval}  
\end{align}
That is, we define proximity as the mean of feature-wise distances between the CF example and the original input. Proximity for a set of examples is simply the average proximity over all the examples. Note that the above metric for continuous proximity  is slightly different than the one used during CF generation.  During CF generation, we transform continuous features to [0, 1] for reasons discussed in Section~\ref{sec:practical}, but we use the features in their original scale during evaluation for better interpretability of the distances.

\para{Sparsity. }
While proximity quantifies the average change between a CF example and the original input, we also measure 
another related property, sparsity, that captures the \emph{number} of features that are different. 
We define sparsity as the number of changes between the original input and a generated counterfactual.
\begin{equation}
    \texttt{Sparsity}: 1-\frac{1}{kd} \sum\limits_{i=1}^{k} \sum\limits_{l=1}^{d} 1_{[\counterfactual_i^l \neq \vecx_i^l]}
\end{equation}
\noindent where $d$ is the number of input features. For clarity, we can also define sparsity separately for continuous and categorical features (where \texttt{Categorical-Proximity} is identical to \texttt{Categorical-Sparsity}).
Note that greater values of sparsity and proximity are desired.

\para{Diversity.}
Diversity of CF examples can be evaluated in an analogous way to proximity. Instead of feature-wise distance from the original input, we measure feature-wise distances between each pair of CF examples, thus providing a different metric for evaluation than the loss formulation from Equation~\ref{eqn:diversity-dpp}.  Diversity for a set of counterfactual examples is the mean of the distances between each pair of examples. Similar to proximity, we compute separate diversity metrics for categorical and continuous features. 
$$ \texttt{Diversity}: \Delta = \frac{1}{\Comb{k}{2}} \sum\limits_{i=1}^{k-1} \sum\limits_{j=i+1}^{k} \operatorname{dist}(\counterfactual_i, \counterfactual_j),$$
\noindent where $\operatorname{dist}$ is either $\operatorname{dist\_cont}$ or $\operatorname{dist\_cat}$.

In addition, we define an analagous sparsity-based diversity metric that measures the fraction of features that are different between any two pair of counterfactual examples. 
$$ \texttt{Count-Diversity} : \frac{1}{C^2_k d} \sum\limits_{i=1}^{k-1} \sum\limits_{j=i+1}^{k} \sum\limits_{l=1}^{d} 1_{[\counterfactual_i^l \neq \counterfactual_j^l]}   $$
It is important to note that the evaluation metrics used here are intentionally 
different from Equation~\ref{eq:diversity}, so there is no guarantee that our generated counterfactuals would do well on all these metrics, especially on the sparsity metric which is not optimized explicitly in CF generation.
In addition, given the trade-off between diversity and proximity, no method will be able to maximize both. Therefore, evaluation of a counterfactual set will depend on the relative merits of diversity versus proximity for a particular application domain.

\subsection{Approximating the local decision boundary}
\label{sec:nn-eval}
The above properties are desirable, but ideally, we would like to evaluate whether the examples help a user in \emph{understanding} the local decision boundary of the ML model.  As a tool for explanation, counterfactual examples help a user intuitively explore specific points on the other side of the ML model's decision boundary, which then help the user to ``guess'' the workings of the model. To construct a metric for the accuracy of such guesses, we approximate a user's guess with \emph{another} machine learning model that is trained on the 
generated counterfactual examples and the original input.  Given this secondary model, we can evaluate the effectiveness of counterfactual examples by comparing how well the secondary model can mimic the original ML model. Thus, considering the secondary model as a best-case scenario of how a user may rationally extrapolate counterfactual examples, we obtain a proxy for how well a user may guess the local decision boundary.

Specifically, given a set of counterfactual examples and the input example, we train a 1-nearest neighbor (\emph{1-NN}) classifier that predicts the output class of any new input. Thus, an instance closer to any of the CF examples will be classified as belonging to the desired counterfactual outcome class, and instances closer to the original input will be classified as the original outcome class.   We chose 1-NN for its simplicity and connections to people's decision-making in the presence of examples. 
We then evaluate the accuracy of this classifier against the original ML model on a dataset of simulated test data. 
To generate the test data, we consider samples of increasing distance from the original input. 
Consistent with training, we scale distance for continuous features by dividing it by the median absolute deviation (MAD) for each feature. 
Then, we construct a hypersphere centered at the original input that has  dimensions equal to the number of continuous features. Within this hypersphere, we sample feature values uniformly at random.  For categorical features,  in the absence of a clear distance metric, we uniformly sample across the range of possible levels.

In our experiments, we consider spheres with radiuses as multiples of the MAD ($r=\{0.5,1,2\}MAD$). For each original input,  we sample 1000 points at random per sphere to evaluate how well the secondary 1-NN model approximates the local decision boundary.  
{\em Note that this 1-NN classifier is trained from a handful of CF examples, and we intentionally choose this simple classifier to approximate what a person could have done given these CF examples.}

\begin{figure*}[t]
    \includegraphics[width=0.88\textwidth]{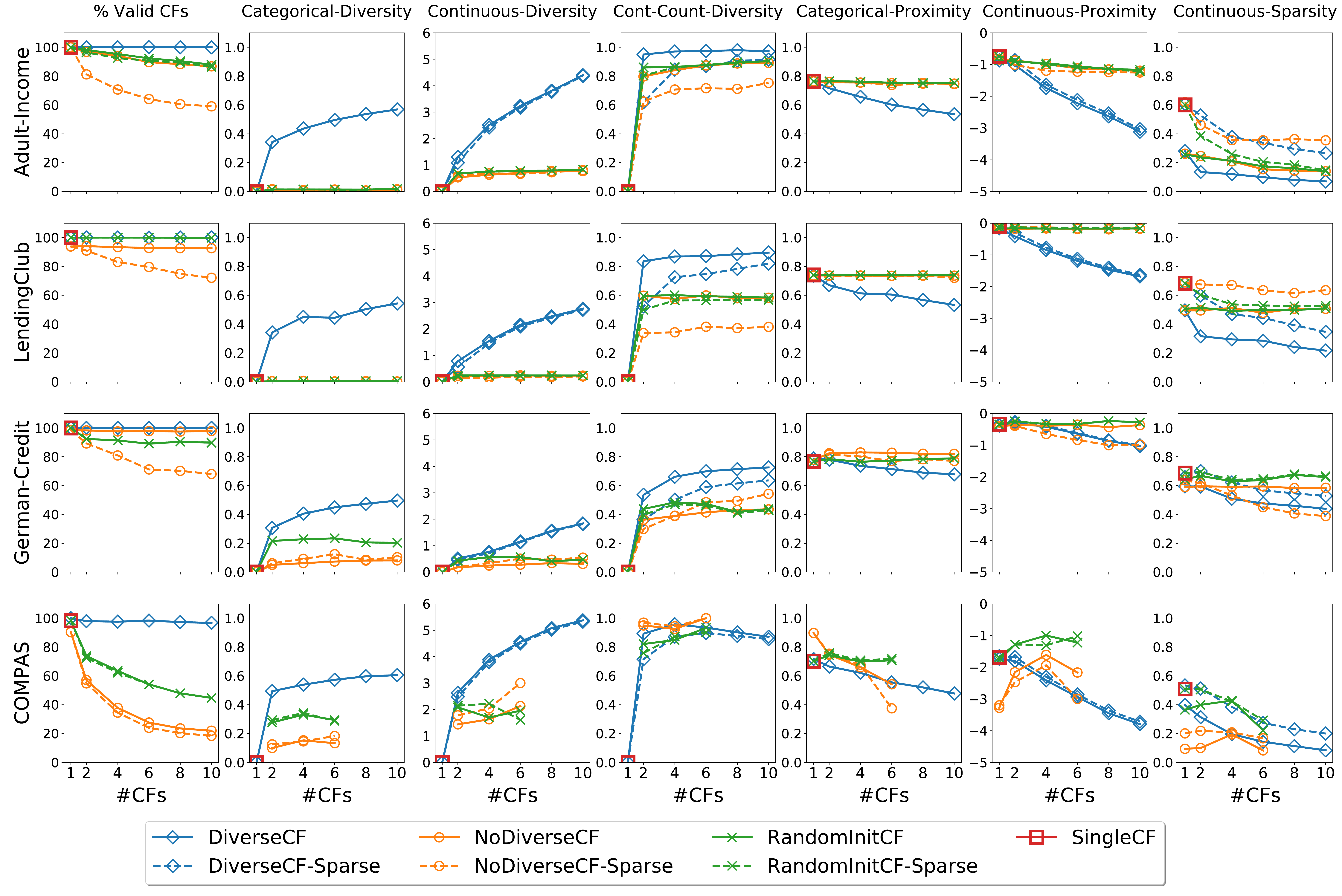}
    \caption{Comparisons of \diverseCF with baseline methods on \%Valid CFs, diversity, proximity and sparsity (\singleCF only shows up for $k=1$).
    All $y$-axes are defined so that higher values are better, while the $x$-axis represents the number of requested counterfactuals.
    \diverseCF finds a greater number of valid unique counterfactuals for each $k$ and tends to generate more diverse counterfactuals than the baseline methods.
    For \compas dataset, none of the baselines could generate $k>6$ CF examples for any original input, therefore we only show results for \diverseCF in COMPAS when $k>6$. 
    }
    \label{fig:compare-loss-metrics}
\end{figure*}

\subsection{Datasets}
\label{sec:data}

To evaluate our method, we consider the following four datasets.

\para{Adult-Income.}
This dataset contains demographic, educational, and other information based on 1994 Census database and is available on the UCI machine learning repository~\cite{adult}. We preprocess the data based on a previous analysis~\cite{adult-cleaning} and obtain $8$ features, namely, hours per week, education level, occupation, work class, race, age, marital status, and sex.  The ML model's task is to classify whether an individual's income is over $\$50,000$. 

\para{LendingClub.}
This dataset contains five years (2007-2011) data on loans given by LendingClub, an online peer-to-peer lending company. We preprocess the data based on previous analyses~\cite{lending1,lending2,tan2017distill} and obtain $8$ features, namely, employment years, annual income, number of open credit accounts, credit history, loan grade as decided by LendingClub, home ownership, purpose, and the state of residence in the United States. The ML model's task is to decide loan decisions based on a prediction of whether an individual will pay back their loan.

\para{German-Credit.}
This dataset contains information about individuals who took a loan from a particular bank~\cite{german}. We use all the 20 features in the data, including several demographic attributes and credit history, without any preprocessing. The ML model's task is to determine whether the person has a good or bad credit risk based on their attributes.

\para{COMPAS.}
This dataset was collected by ProPublica~\cite{propublica-story} as a part of their analysis on recidivism decisions in the United States. We preprocess the data based on previous work~\cite{dressel2018accuracy} and obtain $5$ features, namely, bail applicants' age, gender, race, prior count of offenses, and degree of criminal charge. The ML model's task is to decide bail based on predicting which of the bail applicants will recidivate in the next two years. 

These datasets contain different numbers of continuous and categorical features as shown in Table~\ref{tab:model}. \compas dataset has a single continuous feature, while \adult, \lclub and \german have 2, 4, and 5 continuous features respectively. 
For all three datasets, we transform categorical features by using one-hot-encoding, as described in Section~\ref{sec:cf-engine}. Continuous features are scaled between 0 and 1. 
To obtain an ML model to explain, we divide each dataset into  80\%-20\% train and test sets, and use cross-validation on the train set to optimize hyperparameters.
To facilitate comparisons with \citet{russell2019efficient}, we use  TensorFlow library to train both a linear (logistic regression) classifier and a non-linear neural network model with a single hidden layer. 
Table~\ref{tab:model} shows modelling details and test set accuracy on each dataset.

\begin{table}[t]
\centering
\small{
\begin{tabular}{lrrrr}
\toprule
Dataset & Linear & Non-linear & Num cat & Num cont \\ \hline
Adult-Income & 0.82 & 0.82 & 6 & 2\\
LendingClub & 0.67 & 0.66 & 4 & 4\\
German-Credit & 0.73 & 0.77 & 15 & 5\\
COMPAS & 0.67 & 0.67 & 4 & 1\\
\bottomrule
\end{tabular}}
\caption{Model accuracy and feature information.}
\label{tab:model}
\end{table}

\subsection{Baselines}
We employ the following baselines for generating CF examples. 
\begin{itemize}[leftmargin=*]
    \item  \singleCF : We follow \citet{wachter2017counterfactual} and generate a single CF example, optimizing for y-loss difference and proximity. 
    \item \chrisCF: We use the mixed integer programming  method proposed by \citet{russell2019efficient} for generating diverse counterfactual examples. 
    This method works only for a linear model.
\item \randomCF : Here we extend \singleCF\ to generate $k$ CF examples by initializing the optimizer independently with $k$ random starting points from $[0,1]^d$. Since the optimization loss function is non-convex, one might obtain different CF examples. 
    \item \nodiverseCF : This method utilizes our proposed loss function that optimizes the set of $k$ examples simultaneously (Equation~\ref{eq:diversity}), but ignores the diversity term by  setting $\lambda_2=0$. 

\end{itemize}

To these baselines, we compare our proposed method, \diverseCF, that generates a set of counterfactual examples and optimizes for both diversity and proximity. As with \randomCF, we initialize the optimizer with random starting points. In addition, we consider a variant \diverseCFsp that performs post-hoc sparsity enhancement on continuous features as described in Section~\ref{sec:practical}.  Similarly, 
for \randomCF and \nodiverseCF, we include results both with and without the sparsity correction. For all methods, we use the ADAM optimizer \citep{kingma2014adam} implementation in TensorFlow (learning rate=0.05) to minimize the loss and obtain CF examples.

In addition, we compare \diverseCF to one of the major feature-based local explanation methods, \lime \citep{ribeiro2016should}, on how well it can approximate the decision boundary. We construct a 1-NN classifier for each set of CF examples as described in \secref{sec:nn-eval}. For LIME, we use the 
prediction of the linear model for each input instance as a local approximation of the ML model's decision surface.  
Note that our 1-NN classifiers are based on only $k \leq 10$ counterfactuals, while LIME's linear classifiers are based on 5,000 samples.

%% file: results.tex
\section{Experiment Results}
\label{sec:results}
In this section, we show that our approach generates a set of more diverse counterfactuals than the baselines according to the proposed evaluation metrics.
We further present examples for a qualitative overview and show that the generated counterfactuals can approximate local decision boundaries as well as LIME, an explanation method specifically designed for local approximation.

\begin{figure*}[ht]
    \includegraphics[width=0.88\textwidth]{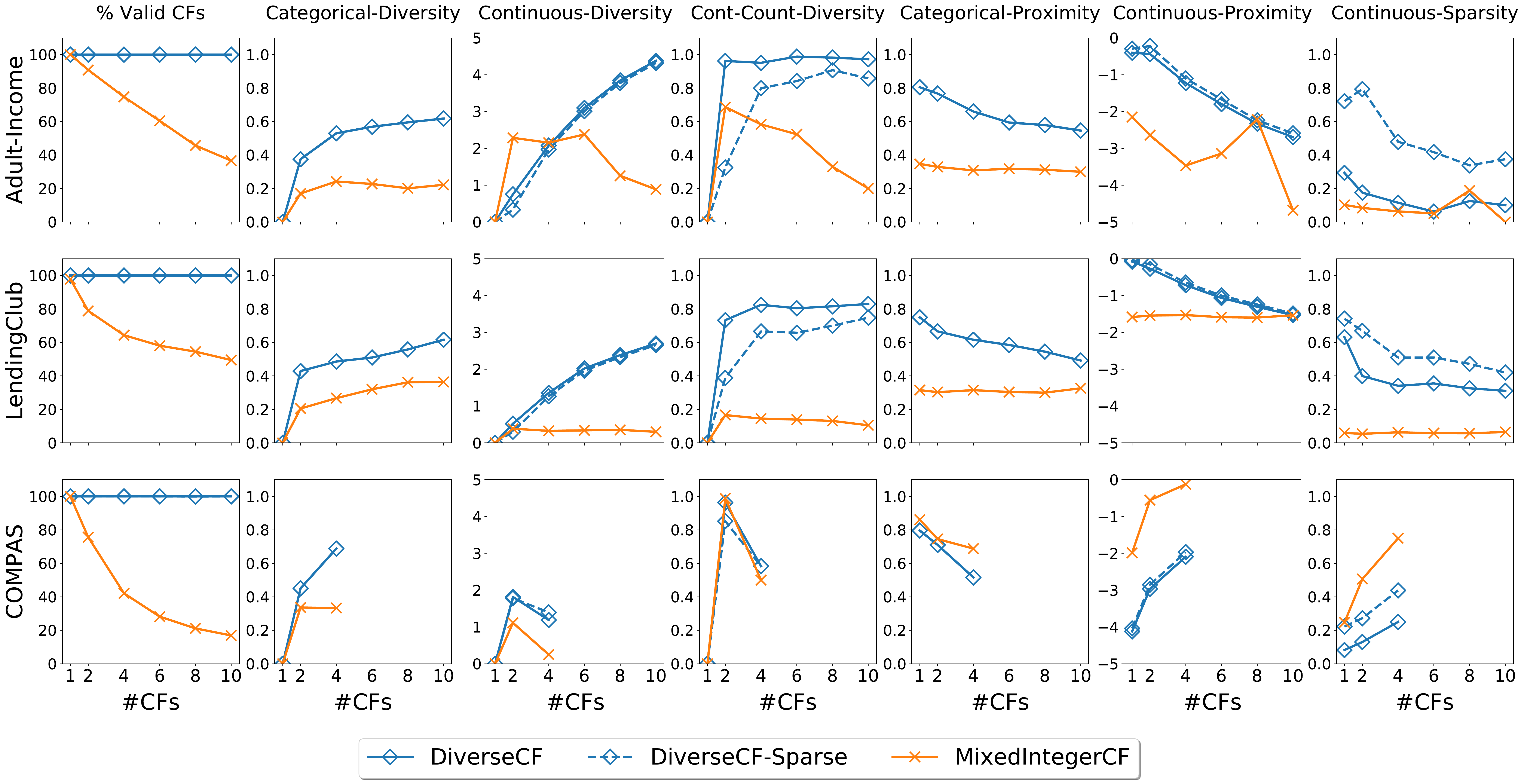}
    \caption{Comparisons of \diverseCF with \chrisCF on \%Valid CFs, diversity, proximity and sparsity on linear ML models.
    For a fair comparison, we compute average metrics only over the original inputs where \chrisCF returned the required number of CF examples. Thus, we omit results when $k>4$ for COMPAS since \chrisCF could not find more than four CFs for any original input. 
    Results for \german are in the Supplementary Materials. 
    }
    \label{fig:compare-loss-metrics-linear}
\end{figure*}

\subsection{Quantitative Evaluation}
\label{sec:quanteval}
We first evaluate \diverseCF based on quantitative metrics of valid CF generation, diversity, and proximity. As described in Section~\ref{sec:cf-engine}, 
 we report results  with hyperparameters, $\lambda_1=0.5$ and $\lambda_2=1$ from Equation~\ref{eq:diversity}.
\figref{fig:compare-loss-metrics} shows the comparison with \singleCF, \randomCF, and \nodiverseCF for explaining the non-linear ML models, while \figref{fig:compare-loss-metrics-linear} compares with \chrisCF for explaining the linear ML models. All results are based on 500 random instances from the test set.

\subsubsection{Explaining a non-linear ML model (\figref{fig:compare-loss-metrics})}
Given that non-linear ML models are common in real world applications, we focus our discussion on explaining non-linear models.

\para{Validity.}
Across all four datasets, we find that \diverseCF\ generates nearly 100\% valid CF examples for all values of the requested number of examples $k$.  Baseline methods without an explicit diversity objective can generate valid CF examples for $k=1$, but their percentage of unique valid CFs decreases as $k$ increases. 
Among the datasets, we find that it is easier to generate valid CF examples for \lclub (\randomCF also achieves $\sim$100\% validity) while  \compas is the hardest, likely driven  by the fact that it has only one continuous feature---prior count of offenses. As an example, at $k=10$ for \compas, a majority of the CFs generated by the next best method, \randomCF are either duplicate or invalid.

\para{Diversity.}
Among the valid CFs, \diverseCF\ also generates more diverse examples than the baseline methods for both continuous and categorical features.
For all datasets, Continuous-Diversity for \diverseCF is the highest and increases as $k$ increases, reaching up to eleven times the baselines for the \lclub dataset at $k=10$. 
Among categorical  features, average number of different features between CF examples is higher for all datasets than baseline methods, especially for \adult and \lclub\ datasets  where \texttt{Cat-Diversity} remains close to zero for baseline methods. 
  Remarkably, \diverseCF has the highest number of continuous features changed too (\texttt{Cont-Count-Diversity}), even though it was not explicitly optimized for this metric. The only exception is on \compas data where \nodiverseCF has a slightly higher \texttt{Cont-Count-Diversity} for $k<=6$, but is unable to generate any valid CFs for higher $k$'s.

\para{Proximity.}
To generate diverse CF examples, \diverseCF\ searches a larger space than proximity-only methods such as \randomCF\ or \nodiverseCF. As a result, 
\diverseCF\  returns examples with lower proximity than other methods, indicating an inherent tradeoff between diversity and proximity. 
However,  for categorical features, the difference in proximity compared to baselines is small, up to $\sim$30\% of the baselines' proximity. 
Higher proximity over continuous features can be obtained by adding the post-hoc sparsity enhancement (\diverseCFsp), which results in higher sparsity than \diverseCF for all datasets (but correspondingly  lower count-based diversity). Thus, this method can be used to fine-tune \diverseCF towards more proximity if desired.

\subsubsection{Explaining linear ML models (\figref{fig:compare-loss-metrics-linear})}
To compare our methods with \chrisCF~\cite{russell2019efficient}, we 
explain a linear ML model for each dataset.
Similar to the results on non-linear models, \diverseCF outperforms \chrisCF by finding 100\% valid counterfactuals, and the gap with \chrisCF increases as $k$ increases.  
We also find that \diverseCF has consistently higher diversity among counterfactuals than \chrisCF for all datasets.
Importantly, better diversity in the CFs from \diverseCF does not come at the price of proximity. For \adult and \lclub datasets, 
\diverseCF has better proximity and sparsity than \chrisCF.

\input{qual_table}
\subsection{Qualitative evaluation}\label{sec:qual-results}

To understand more about the resultant explanations, we look at 
sample CF examples generated by \diverseCF with sparsity in Table~\ref{tab:qual-examples}. 
In the three datasets,\footnote{We skip \german for space reasons.} the examples capture some intuitive variables and vary them: Education in \adult, Income in \lclub dataset, and PriorsCount in \compas. In addition, the user also sees other features that can be varied for the desired outcome. For example, in the \compas\ input instance, 
a person would have been granted bail if they had been a 
Caucasian or charged with Misdemeanor instead of Felony. 
These features do not really lead to actionable insights because the subject cannot easily change them, 
but nevertheless provide the user an accurate picture of scenarios where they would have been out on bail (and also raise questions about potential racial bias in the ML model itself). 
In practice, we expect that a  domain expert or the user may provide \emph{unmodifiable} features which \diverseCF can treat as constants in the counterfactual generation process. 

Similarly, in the \adult dataset, the set of counterfactuals show that studying for an advanced degree can lead to a higher income, but also shows less obvious counterfactuals such as getting married for a higher income (in addition to finishing professional school and increasing hours worked per week). These counterfactuals are likely generated due to underlying correlations in the dataset (married people having higher income). To counter such correlational outcomes and preserve known causal relationships, we present a post-hoc filtering method in Section~\ref{sec:causal}. 

These qualitative examples also confirm our observation regarding sparsity and the choice of the $yloss$ function. Continuous variables in counterfactual examples (e.g., income in \lclub) never change to their maximum extreme values thanks to the hinge loss, which was an issue using other $yloss$ metrics such as $\ell_1$ loss. Furthermore,   
it does not require changing a large number of features to achieve the desired outcome.
However, based on the domain and use case, a user may prioritize changing certain variables or desire more sparse or more diverse CF examples. As we described in Section~\ref{sec:practical}, these variations can be achieved by appropriately tuning weights on features and the learning rate for optimization.

Overall, these initial set of CF examples help understand the important variations as learned by the algorithm. We expect the user to engage their actionability constraints with this initial set to iteratively generate focused CF examples, that can help find useful variations. In addition, these examples can also expose biases or odd edge-cases in the ML model,
which can be useful for model builders in debugging, or for fairness evaluators in discovering 
bias.

\begin{figure*}[t]
    \centering
    \begin{subfigure}{0.39\textwidth}
        \centering
        \includegraphics[width=0.95\textwidth]{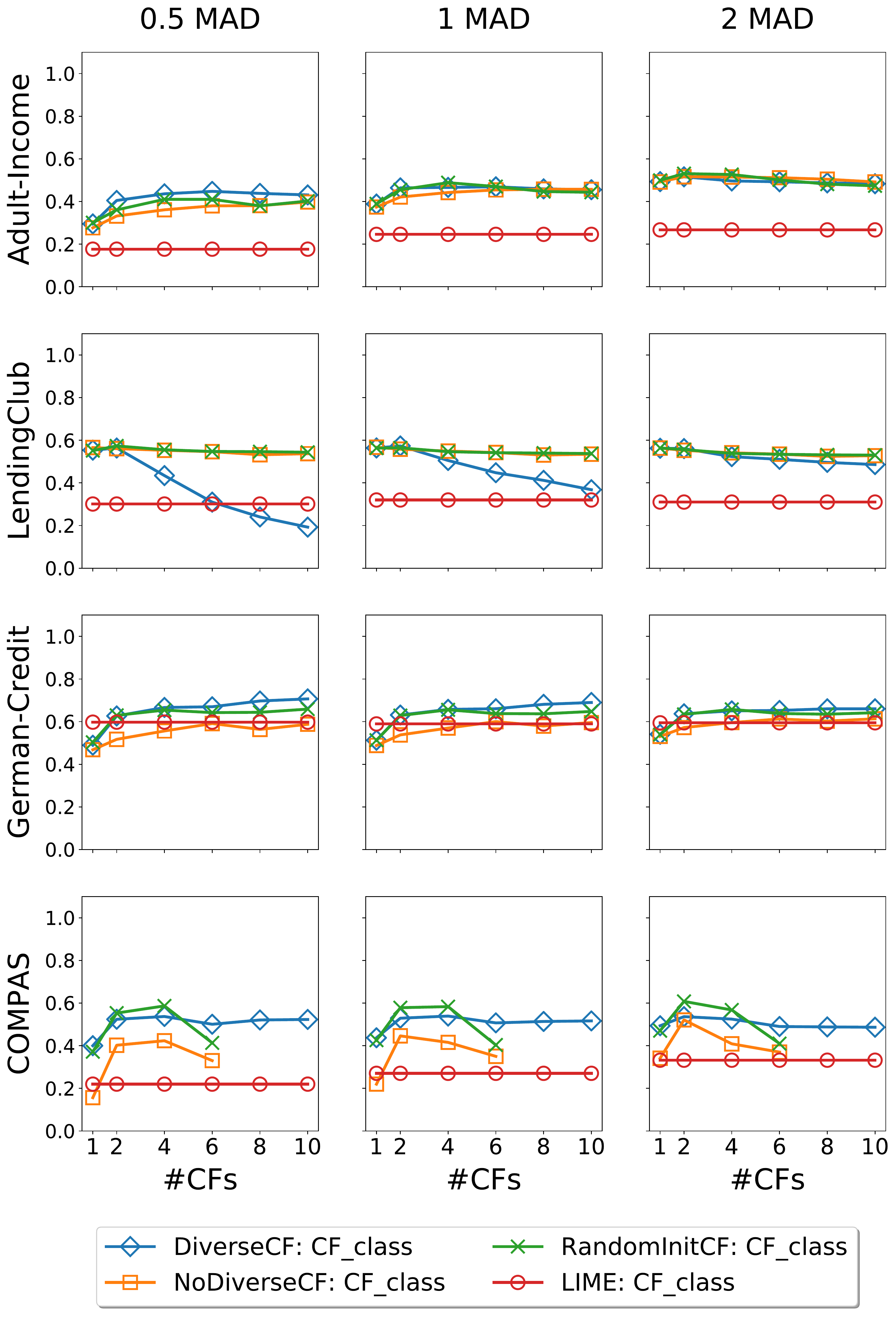}
        \caption{F1 score of the counterfactual class.}
        \label{fig:f1}
    \end{subfigure}
    \hfill
    \begin{subfigure}{0.29\textwidth}
        \centering
        \includegraphics[width=0.9\textwidth]{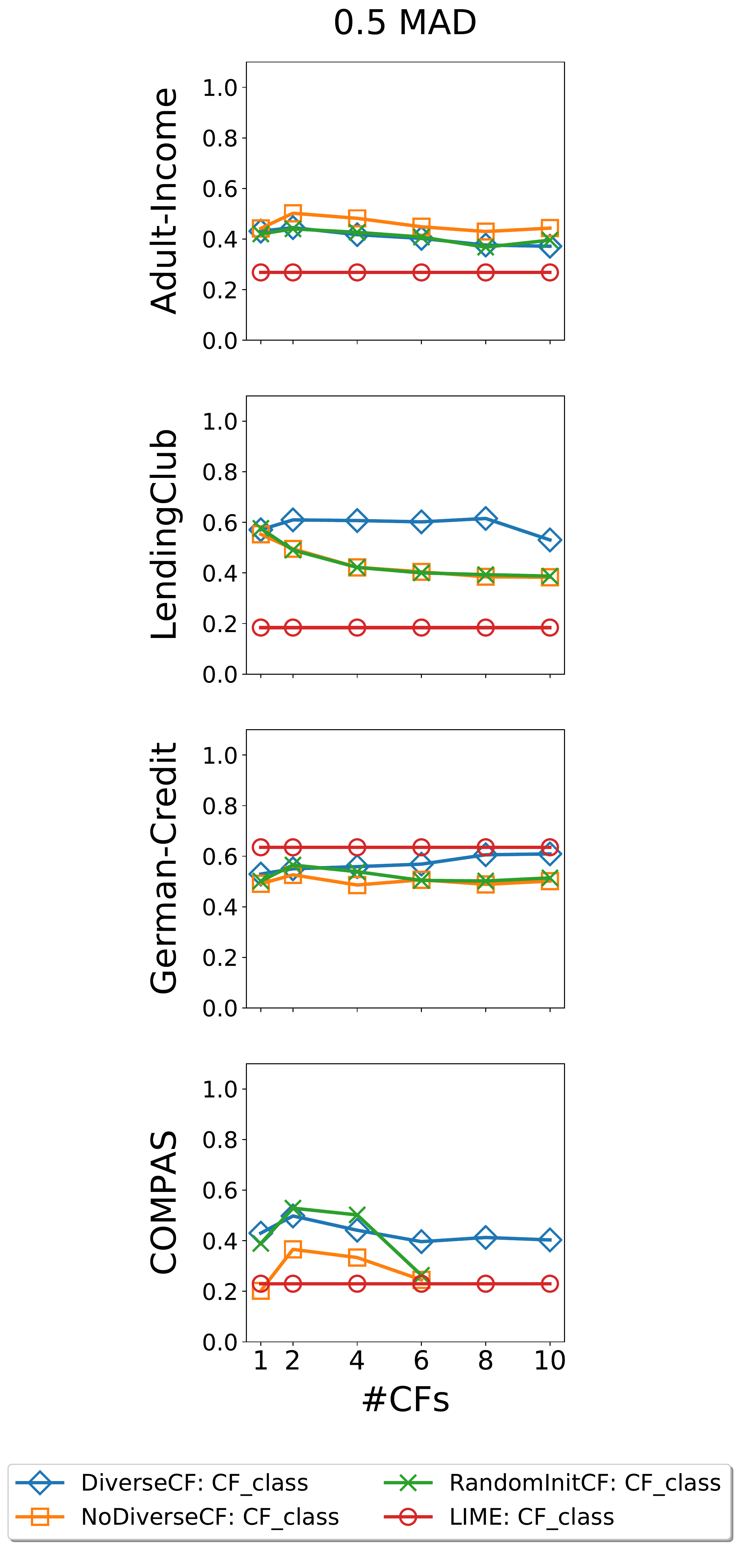}
        \caption{Precision.}
        \label{fig:precision}
    \end{subfigure}
    \begin{subfigure}{0.29\textwidth}
        \centering
        \includegraphics[width=0.9\textwidth]{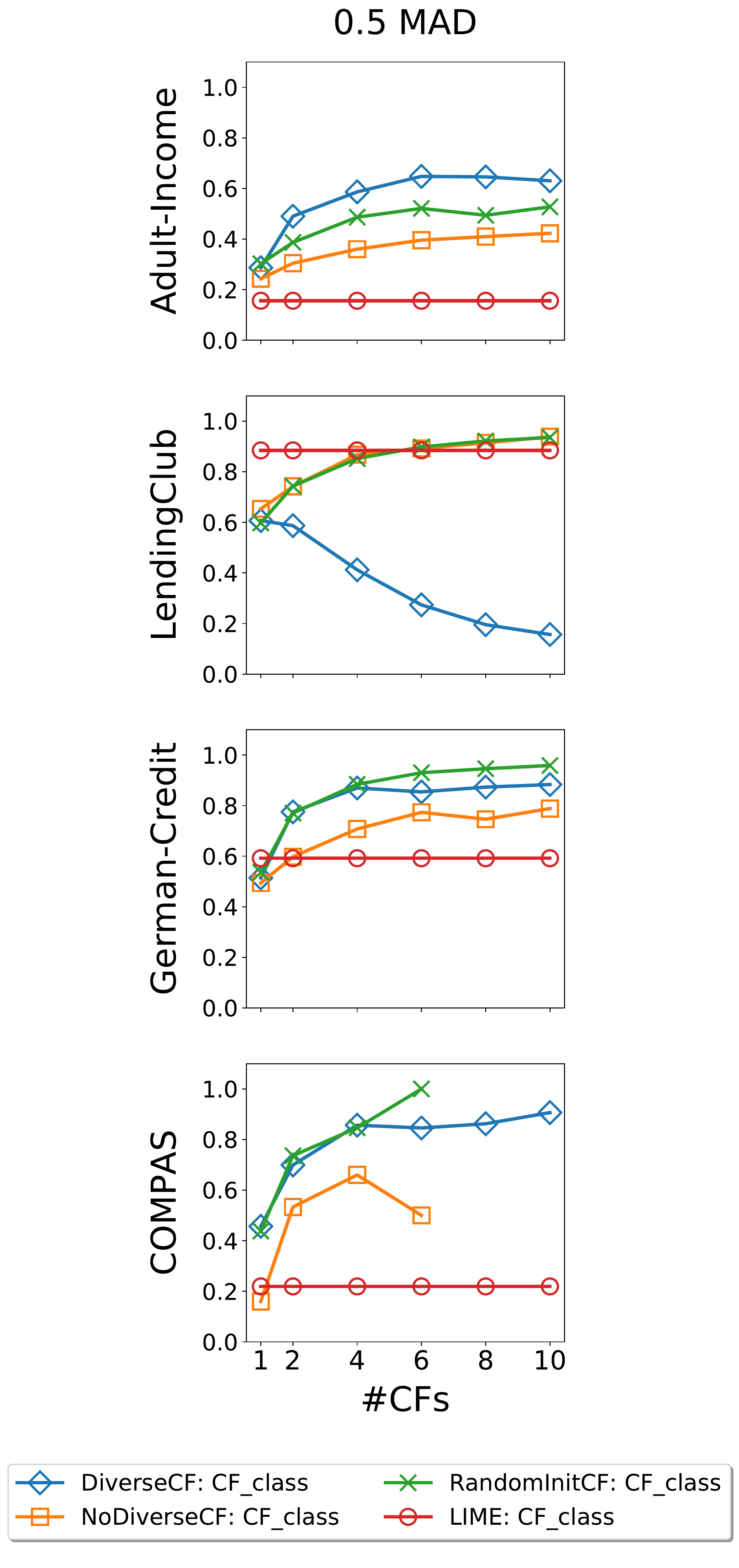}
        \caption{Recall.}
        \label{fig:recall}
    \end{subfigure}
    \caption{Performance of 1-NN classifiers learned from counterfactuals at different distances from the original input.
    \diverseCF  outperforms LIME and baseline CF methods  in F1 score on correctly predicting the counterfactual class, except in \lclub dataset. For \adult and \compas datasets, both precision and recall is higher for \diverseCF compared to \lime.
    }
    \label{fig:f1-score}
\end{figure*}

\subsection{Approximating local decision boundary}
As a proxy for understanding how well users can guess the local decision boundary of the ML model (see Section~\ref{sec:nn-eval}), we compare classifiers based on the proposed \diverseCF\ method, baseline methods, and \lime. 
We use precision, recall, and F1 for the counterfactual outcome class (\figref{fig:f1-score}) as our main evaluation metric because of the class imbalance in data points near the original input. To evaluate the sensitivity of these metrics to varying distance from the original input, we show these metrics for points sampled within 
varying distance thresholds.

Even with a handful (2-11) of training examples (generated counterfactuals and the original input), we find that 1-NN classifiers trained on the output of \diverseCF obtain 
higher F1 score than the LIME classifier in most configurations.  
For instance, on the \adult dataset, at $k=4$ and 0.5MAD threshold,  \diverseCF obtains $F1=0.44$ while \lime obtains $F1=0.19$. This result stays consistent as we increase $k$ or the distance from the original input.  
One exception is 
\lclub at 0.5 MAD and $k=10$ where the F1 score for \diverseCF drops below \lime. Figures 3b and 3c indicate that this drop is due to low recall for \diverseCF at this configuration. Still, precision remains substantially higher for \diverseCF ($0.61$) compared to $0.19$ for \lime. This observation is likely because \lime predicts a majority of the instances 
as the CF class for this dataset, whereas \diverseCF has fewer false positives.  On \adult and \compas datasets, \diverseCF achieves both higher precision and recall than \lime.

As for the difference between different methods of generating counterfactuals,
\diverseCF tend to perform similarly with \nodiverseCF and \randomCF in terms of F1, except in \lclub.
An advantage of \diverseCF is that it can handle high values of $k$ for which \nodiverseCF and \randomCF cannot find $k$ unique and valid counterfactuals.
Another intriguing observation is that the performance improves very quickly as the number of counterfactuals ($k$) increases, which suggests that two counterfactuals may be sufficient for a 
1-NN classifier to get a reasonable idea of the data distribution around $\vecx$ in these four datasets. 
This observation may also be 
why \diverseCF provides similar F1 score compared to baselines, and merits further study on more complex datasets.

Overall, these results show that examples from \diverseCF  can   approximate the local decision boundary 
at least as well as local explanation methods like LIME. 
Still, the gold-standard test will be to conduct a behavioral study where people evaluate whether CF examples provide better explanation than past approaches, which we leave for future work.

%% file: qual_table.tex
\begin{table*}[t]
\footnotesize
\begin{tabular}{l|llllllll}
\toprule
\textbf{Adult}                                                                      & HrsWk & Education   & Occupation   & WorkClass     & Race  & AgeYrs & MaritalStat & Sex    \\ \hline
\begin{tabular}[c]{@{}l@{}}Original input\\ (outcome: \textless{}=50K)\end{tabular} & 45.0  & HS-grad     & Service      & Private       & White & 22.0   & Single      & Female \\ \hline
\textbf{}   & ---  & Masters & ---        & --- & --- & 65.0   & Married   & Male \\
Counterfactuals       
& ---  & Doctorate & --- & Self-Employed       & --- & 34.0   & ---     & --- \\
(outcome: \textgreater{}50K)                                                        & 33.0  & ---     & White-Collar      & ---       & --- & 47.0   & Married   & ---   \\  & 57.0  & Prof-school   & --- & ---       & --- & ---  & Married      & --- 
\\ \bottomrule
\end{tabular}

\begin{tabular}{l|llllllll}
\toprule
\textbf{LendingClub} & {EmpYrs} & {Inc\$} & {\#Ac} & {CrYrs} & {LoanGrade} & {HomeOwner} & {Purpose} & {State} \\ \hline
{\begin{tabular}[c]{@{}l@{}}Original input\\ (outcome: Default)\end{tabular}} & 7.0             & 69996.0        & 4.0            & 26.0           & D                  & Mortgage           & Debt             & NY             \\ \hline & ---             & 61477.0        & ---            & ---           & B                  & ---                & Purchase         & ---             \\
{Counterfactuals}                                                             & 10.0             & 83280.0       & 1.0            & 23.0            & A                  & ---           & ---             & TX             \\
{(outcome: Paid)}                                                             & 10.0             & 69798.0       & ---            & 40.0            & A                  & ---               & ---             & ---             \\& 10.0            & 130572.0       & ---           & ---           & A                  & Rent           & ---             & ---          \\ \bottomrule
\end{tabular}
\begin{tabular}{l|lllll}
\toprule
\textbf{COMPAS}                                                                     & PriorsCount & CrimeDegree & Race             & Age              & Sex    \\ \hline
\begin{tabular}[c]{@{}l@{}}Original input\\ (outcome: Will Recidivate)\end{tabular} & 10.0        & Felony      & African-American & \textgreater{}45 & Female   \\ \hline
 & ---        & ---      & Caucasian & --- & --- \\
Counterfactuals                                                                     & 0.0         & ---      & ---  & ---          & Male   \\
(outcome: Won't Recidivate)                                                         & 0.0         & --- & Hispanic  & --- & ---   \\
\textbf{}                                                                           & 9.0        & Misdemeanor & ---            & ---          & ---   \\ \bottomrule
\end{tabular}
\caption{Examples of generated counterfactuals in \adult, \lclub and \compas datasets.
}\label{tab:qual-examples}
\end{table*}

%% file: causal.tex
\section{Causal feasibility of CF Examples}
\label{sec:causal}

\begin{figure}
    \includegraphics[scale=0.28]{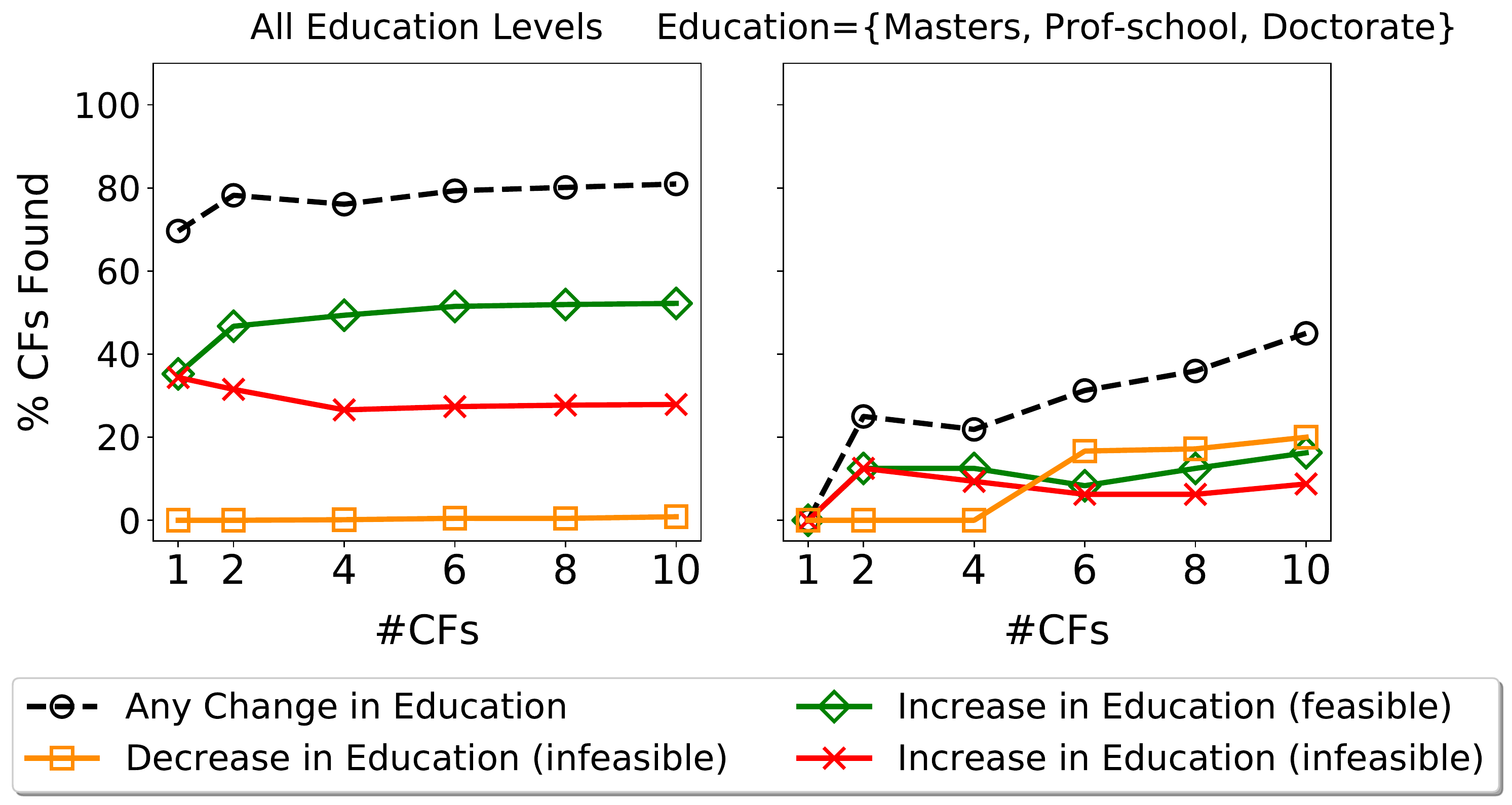}
    \caption{Post-hoc filtering of CF examples based on  causal constraints.
        The left figure shows that there are nearly 80\% CFs that include any change in education, out of which more than one-third 
        are infeasible. If we filter to only people with higher degrees, 
    almost half of the changes in educational degrees are infeasible. 
    }
    \label{fig:causal}
\end{figure}

So far, we have generated CF examples by varying each feature independently. However, this can  lead to infeasible examples, since many features are causally associated with each other.
For example, in the loan application, it can be almost impossible for a person to 
obtain a higher educational degree without spending time (aging).
Consequently, while being valid, diverse and proximal, such a CF example is not feasible and thus not actionable by the person. In this context, we argue that incorporating causal models of data generation is important to 
prevent infeasible counterfactuals.

Here we present a simple way of incorporating causal knowledge in our proposed method. Users can provide their domain knowledge in the form of pairs of features and the direction of the causal edge between them~\cite{pearl2009causality}. Using this, we construct constraints that any counterfactual should follow. For instance, any counterfactual that changes the cause without  changing its outcome is infeasible. Given these constraints, we apply a filtering step after CF examples are generated, to increase the feasibility of the output CF set. 

As an example, we consider two infeasible changes based on the causal relationship between educational level and age,  $\texttt{Education} \uparrow \ \Rightarrow \texttt{Age} \uparrow$, 
and on the practical constraint that educational level of a person cannot be decreased, $\texttt{Education} \cancel{\downarrow}$.
As \figref{fig:causal} shows, 
over one-third of the obtained counterfactuals that include a change in education level are infeasible and need to be filtered: most of them suggest an individual to obtain a higher degree but do not increase their age. 
In fact, this fraction increases as we look at CF examples for  highly educated people (Masters, Doctorate and Professional):  
as high as 50\% of all CFs suggest to switch to a lower education degree. 
Though post-hoc filtering can ensure feasibility of the resultant CF examples, 
it is more efficient to incorporate causal constraints during  CF generation.
We leave this for future work.

%% file: conclusion.tex
\section{Concluding Discussion}
\label{sec:conclusion}

Building upon prior work on counterfactual explanations~\cite{wachter2017counterfactual,russell2019efficient}, we proposed a framework for generating and evaluating a diverse and feasible set of counterfactual explanations. 
We demonstrated the benefits of our method compared to past approaches on being able to generate a high number of unique, valid, and diverse counterfactuals for a given input for any machine learning model. 

Here we note directions for future work. First, our method assumes knowledge of the gradient of the ML model. It is useful to construct methods that can work for fully \emph{black-box} ML models.   Second, we would like to incorporate causal knowledge \emph{during} the generation of CF examples, rather than as a post-hoc filtering step.    
Third,  as we saw in \S\ref{sec:qual-results},  it is important to understand people's preferences with respect to what additional constraints to add to our framework.
Providing an intuitive interface to select scales of features and add constraints, and conducting behavioral experiments to support interactive explorations can greatly enhance the value of CF explanation. It will also be interesting to study the tradeoff between diversity and the cognitive cost of making a choice (``choice overload''~\citep{scheibehenne2010can,bundorf2010choice}), as the number of CF explanations is increased.     
Finally, while we focused on the utility for an end-user who is the subject of a ML-based decision,  we argue that CF explanations can be useful for different stakeholders in the decision making process~\cite{tomsett2018interpretable}, including model designers, decision-makers such as a judge or a doctor, and decision evaluators such as 
auditors.

%% file: supp.tex
\newpage
\appendix

\begin{table*}[h!]
\begin{tabular}{l|llll}
\toprule
                                                                                                       & COMPAS           & \begin{tabular}[c]{@{}l@{}}Adult \\ Income\end{tabular}             & \begin{tabular}[c]{@{}l@{}}German \\ Credit\end{tabular} & \begin{tabular}[c]{@{}l@{}}Lending \\ Club\end{tabular}       \\ \hline
\begin{tabular}[c]{@{}l@{}}\# Continuous \\ Features\end{tabular}                                      & 1                & 2      & 5           & 4                                                             \\ \hline
\begin{tabular}[c]{@{}l@{}}\# Categorical \\ Features\end{tabular}                                     & 4                & 6      & 15           & 4                                                             \\ \hline
\begin{tabular}[c]{@{}l@{}}Range across all \\ Continuous Features\\ (Min, Avg, Max)\end{tabular}      & (0, 3.5, 38)    & (1, 39.5, 99)     & \begin{tabular}[c]{@{}l@{}}(1, 668, \\ 15945)\end{tabular} & \begin{tabular}[c]{@{}l@{}}(1, 16292, \\ 200000)\end{tabular} \\ \hline

\begin{tabular}[c]{@{}l@{}}\# Levels across all \\ Categorical Features\\ (Min, Avg, Max)\end{tabular} & (2, 3.25, 6)     & (2, 4.5, 8)       &  (2, 4, 10) & (4, 5.5, 7)                                                   \\ \hline
Undesired Class                                                                                        & Will Recidivate  & \textless{}=50K  &  Bad & Default                                                       \\ \hline
\begin{tabular}[c]{@{}l@{}}Desired \\ Counterfactual Class\end{tabular}                                & Won't Recidivate & \textgreater{}50K & Good & Paid                                                          \\ \hline
Training Data Size                                                                                     & 1443             & 6513       & 800       & 8133                                                          \\ \hline
\begin{tabular}[c]{@{}l@{}}Fraction of Instances\\ with Desired CF\\  Outcome\end{tabular}             & 0.55       & 0.30      & 0.25              & 0.8                                                           \\ \hline
Nonlinear Model                                                                                          & ANN(1, 20)       & ANN(1, 20)      & ANN(1, 20)   & ANN(1, 5)                                                     \\ \hline
Test set accuracy                                                                                      & 67\%             & 82\%     & 77\%          & 66\%                                                          \\ \bottomrule
\end{tabular}
\caption{Dataset description.}
\label{tab:datasets}
\end{table*}
\section{Supplementary Materials}
\begin{figure*}[ht]
    \includegraphics[width=1.0\textwidth]{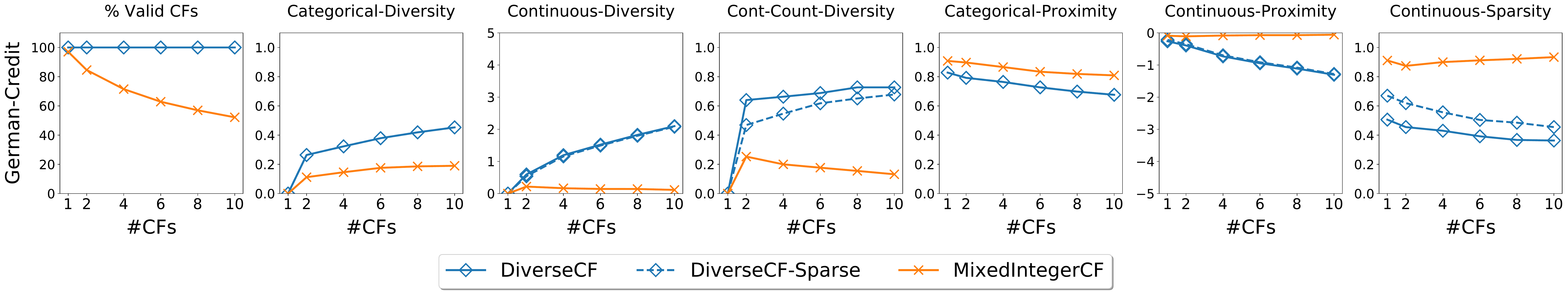}
    \caption{Comparisons of \diverseCF with \chrisCF on \%Valid CFs, diversity, proximity and sparsity on linear ML models for \german.
    For a fair comparison, we compute average metrics only over the original inputs where \chrisCF returned the required number of CF examples.  
    }
    \label{fig:linear-german}
\end{figure*}

Here we discuss the data properties and the implementation details of ML models relevant for reproducing our results. Our open-source implementation is available at {\color{blue} \url{https://github.com/microsoft/DiCE}} which can be used to generate counterfactual examples for any other dataset or ML model. Further, Figure \ref{fig:linear-german} below compares \diverseCF with \chrisCF for explaining the linear ML model over the \german dataset.

\subsection{Building ML Models}

First, we build a ML model for each dataset that gives accuracy comparable to previously established benchmarks, using the Adam Optimizer~\cite{kingma2014adam} in TensorFlow. We described the datasets that we used in our analysis in Section \ref{sec:data}. Table \ref{tab:datasets} provides detailed  properties of the processed data and the ML models that we used. We tuned the hyperparameters of the ML model  based on previous analyses and found that a single hidden layer neural network gives best generalization ability for all datasets. While $20$ hidden neurons worked well for \compas, \adult and \german datasets, increasing more than $5$ neurons worsened the generalization for \lclub dataset. Furthermore, to handle the class imbalance problem while training with these datasets, we oversampled the training instances belonging to the minority class.

\subsection{Explaining linear ML models: \german}
Similar to results we obtained for other datasets, we observe that for \german data, \diverseCF  consistently generates more diverse counterfactuals compared to \chrisCF.

